\documentclass{article}

 \usepackage[preprint]{neurips_2025}

\usepackage[utf8]{inputenc} 
\usepackage[T1]{fontenc}    
\usepackage{hyperref}       
\usepackage{url}            
\usepackage{booktabs}       
\usepackage{amsfonts}       
\usepackage{nicefrac}       
\usepackage{microtype}      
\usepackage{xcolor}         
\usepackage{amsmath}
\usepackage{graphicx} 

\usepackage{algorithm}
\usepackage{algpseudocode}
\usepackage{listings}
\usepackage{xcolor}

\usepackage{booktabs}
\usepackage{multirow}
\usepackage{siunitx}
\usepackage{arydshln} 

\definecolor{projgreen}{RGB}{0,128,128}
\definecolor{projblue}{RGB}{0,0,200}
\lstdefinestyle{ptpseudo}{
  language=Python,
  basicstyle=\ttfamily\small,
  columns=fullflexible,
  keepspaces=true,
  showstringspaces=false,
  breaklines=true,
  frame=single,
  commentstyle=\color{projgreen},
  moredelim=**[is][\color{projblue}]{@@}{@@}
}

\title{Exclusive Self Attention}

\author{%
  Shuangfei Zhai\\
  Apple \\
  \texttt{szhai@apple.com} \\
}

\begin{document}

\maketitle

\begin{abstract}
We introduce exclusive self attention (XSA), a simple modification of self attention (SA) that improves Transformer's sequence modeling performance. The key idea is to constrain attention to capture only information orthogonal to the token's own value vector (thus excluding information of self position), encouraging better context modeling. Evaluated on the standard language modeling task, XSA consistently outperforms SA across model sizes up to 2.7B parameters and shows increasingly larger gains as sequence length grows.
\end{abstract}

\section{Introduction}
Transformers~\citep{vaswani2017attention} consist of interleaved self attention (SA) and feed forward (FFN) layers, where SA aggregates information from the context, and FFN performs position wise feature updates. This SA/FFN design has remarkably stood the test of time and is continuing to serve as the building block for modern Transformer variants. 

In this work, we hypothesize that it is beneficial to further promote the division of labor between SA and FFN. We first expose a peculiar behavior of Transformers, where the output of attention tends to have a high cosine similarity with the self value vector -- and we call it the \textit{\textbf{attention similarity bias}}.

The prevalence of the \textit{attention similarity bias} suggests that SA spends a significant portion of its capacity modeling the point wise feature transformation. This is on the one hand unnecessary, because the information of the current position has a direction residual path to the following FFN layer; and on the other hand harmful, because it creates a competition between modeling the contextual vs point-wise feature. This reasoning directly motivates our solution: exclusive self attention (XSA), which explicitly excludes directions from the attention's output along that of the self value vector. 

We evaluate XSA against standard Transformers on the language modeling task, and we show that XSA 1) introduces minimal computational overhead; 2) achieves better training/validation loss across three model sizes; 3) achieves better downstream evaluation results; 4) maintains consistent gains across different learning rates; 5) shows larger gains as sequence length increases; and 6) is robust w.r.t. the use of attention sinks.


\begin{figure}[!h]
    \centering
    \includegraphics[width=\linewidth]{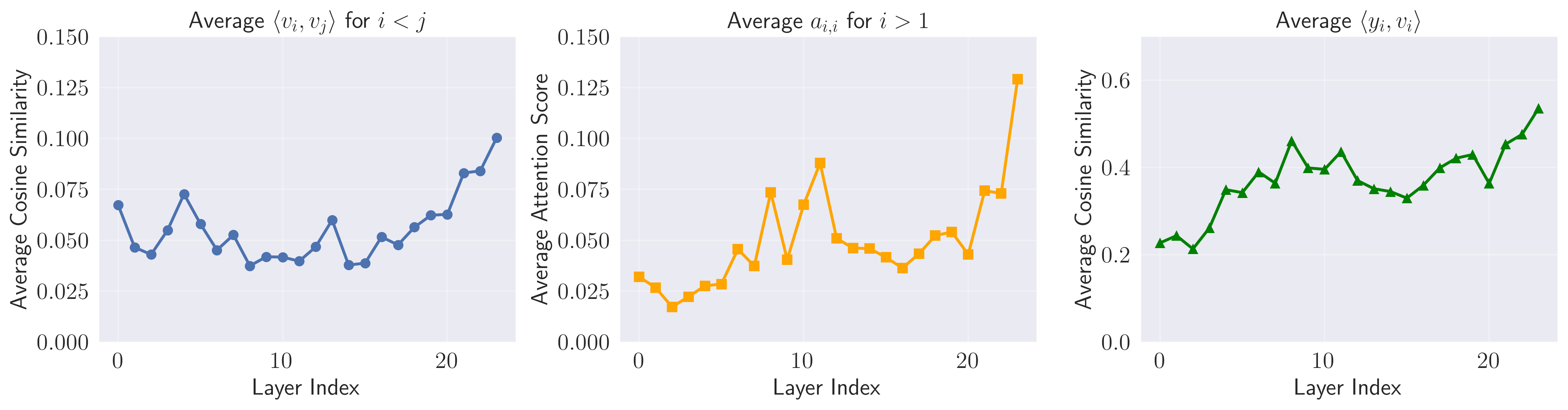}
    \caption{Visualization of the \textit{\textbf{attention similarity bias}} of a 1.3B parameter language model of sequence length 2048 trained for 100B tokens, aggregated on 1024 random training sequences. \textbf{Left}: the average cosine similarity of value vectors $v_i$, $v_j$ within a sequence; \textbf{middle}: the average diagonal attention value $a_{i,j}$; \textbf{right}: the average cosine similarity of attention output $y_i$ and the self value vector $v_i$. See Eq.~\ref{eq:sa} for notations.}
    \label{fig:attn_stats}
\end{figure}

\section{Motivation}
We first define a standard (causal) self attention (SA) as  $y = f(x)$:
\begin{equation}
\label{eq:sa}
    \begin{split}
        &q_i = W_q x_i,\;
        k_j = W_k x_j,\;
        v_j = W_v x_j, \; 
        a_{i,j} = \frac{\exp(q_i^Tk_j)}{\sum_{j'=1}^{i}\exp(q_i^Tk_{j'})},\;
        y_i = \sum_{j=1}^i a_{i,j}v_j,
    \end{split}
\end{equation}
where $W_q, W_k, W_v$ are the query, key and value projections, respectively. 

We next demonstrate the \textit{attention similarity bias} phenomenon, which exposes a hidden problem of SA. To see this, we take a trained language model (which has 1.3B parameters and 2048 sequence length, see Sect.~\ref{sec:experiments} for details) and analyze each of its attention layers in Figure~\ref{fig:attn_stats}. Specifically, we plot three quantities: 1) the average cosine similarity of value vectors $<v_i, v_j>$ within the same head and sequence; 2) the average diagonal values of the attention $a_{i, i}$; and 3) the average cosine similarity of the aggregated values $y_i$ with the corresponding value $v_i$. All quantities are then averaged across attention heads and 1024 random training sequences and plotted for each layer. We can see that 1) value vectors tend to be positively correlated and 2) attention scores to the current position are relatively high. As a direct consequence, there is a high average $<y_i, v_i>$, with an increasing trend w.r.t. the layer index. This suggests that standard SA tends to aggregate value vectors similar to what the self value $v_i$ already encodes, which implicitly overlaps with role of FFN and consequently diminishes SA's goal of context modeling. 

\section{Method}
We define exclusive self attention (XSA) as $z = f(x)$:
\begin{equation}
\label{eq:xsa}
    \begin{split}
        &q_i = W_q x_i,\;
        k_j = W_k x_j,\;
        v_j = W_v x_j, \; 
        a_{i,j} = \frac{\exp(q_i^Tk_j)}{\sum_{j'=1}^{i}\exp(q_i^Tk_{j'})},\;
        y_i = \sum_{j=1}^i a_{i,j}v_j,\\
        &\textcolor{blue}{z_i = y_i - (y_i^Tv_i)\frac{v_i}{\|v_i\|^2_2}}. \;
    \end{split}
\end{equation}
Note that the first line of Equation~\ref{eq:xsa} corresponds to standard SA in Equation~\ref{eq:sa}; XSA introduces an additional step to remove the projection of SA's output $y_i$ on self value vector $v_i$. Consequently, XSA's output $z_i$ no longer contains $v_i$ itself, nor components from that context that's correlated with $v_i$. Therefore, XSA completely removes the \textit{attention similarity bias}.

Our core hypothesis is that, in the presence of residual connections and the FFN block, XSA 1) maintains the expressiveness of standard SA and 2) promotes modeling efficiency by allowing the attention layer to exclusively focus on contextual information. While it is also possible to provide theoretical groundings for it, in this work we defer to empirical evaluations as the main justification.

XSA can be implemented with two lines of code change on top of SA, as demonstrated in Algorithm~\ref{alg:xsa}.

\begin{algorithm}[t]
\caption{PyTorch-style pseudocode for multi-head causal XSA}
\label{alg:xsa}
\small
\begin{lstlisting}[style=ptpseudo]
# x: (B,T,D)
# Wq, Wk, Wv, Wo: (D,D)
# H: number of heads

def exclusive_self_attention(x, Wq, Wk, Wv, Wo, H):
    B, T, D = x.shape

    # linear projections
    Q = (x @ Wq).reshape(B, T, H,  D // H).transpose(1, 2)
    K = (x @ Wk).reshape(B, T, H,  D // H).transpose(1, 2)
    V = (x @ Wv).reshape(B, T, H,  D // H).transpose(1, 2)

    # standard multi-head attention
    Y = torch.nn.functional.scaled_dot_product_attention(Q, K, V, is_causal=True)

    # XSA mode
    @@Vn = torch.nn.functional.normalize(V, dim=-1)
    Z = Y - (Y * Vn).sum(dim=-1, keepdim=True) * Vn
    @@
    # output projection
    out = Z.transpose(1, 2).reshape(B, T, D) @ Wo
    return out
\end{lstlisting}
\end{algorithm}

\section{Experiments}
\label{sec:experiments}
\subsection{Setup}
\paragraph{Codebase} We conduct all experiments with the NanoGPT~\footnote{https://github.com/karpathy/nanoGPT} codebase due to its ease of reproducibility. A few changes are made to the Transformer implementation: we replaced the learned position embeddings with RoPE~\citep{su2023enhanced} whish is a common practice in modern language models; we insert an additional LayerNorm~\citep{ba2016layer} right after the token embeddings which is found to improve training stability; we allow the number of attention heads and the head dimension to be configured independently to allow for more flexible model sizes.

\paragraph{Dataset} We use the FineWeb-100BT~\citep{penedo2024the}~\footnote{https://huggingface.co/datasets/HuggingFaceFW/fineweb} dataset which contains $\sim 100$ billion tokens. We preprocess the dataset following the NanoGPT protocol: namely tokenizing it with the GPT2~\citep{radford2019language} tokenizer and randomly splitting $0.05\%$ of the tokens as the validation set. 

\paragraph{Training details} We closely follow the default training settings of NanoGPT, with a few modifications detailed below. We adopt a default context length of 2048, a global batch size of 256, and a training duration of 200K iterations. This amounts to 100B training tokens which is roughly one epoch over the training set. AdamW~\citep{loshchilov2017decoupled} is used with a linear learning rate warmup of 2K steps, followed by a cosine decay schedule to $\frac{1}{10}$x of the max learning rate. We perform a grid search of learning rate for each baseline model configuration and use it for the XSA variants. We experiment with 3 model sizes, ranging in 0.7B, 1.4B and 2.7B non-embedding parameters. The model configurations are specified in Table~\ref{tab:model_sizes}.

\begin{figure}[!h]
    \centering
    \includegraphics[width=0.9\linewidth]{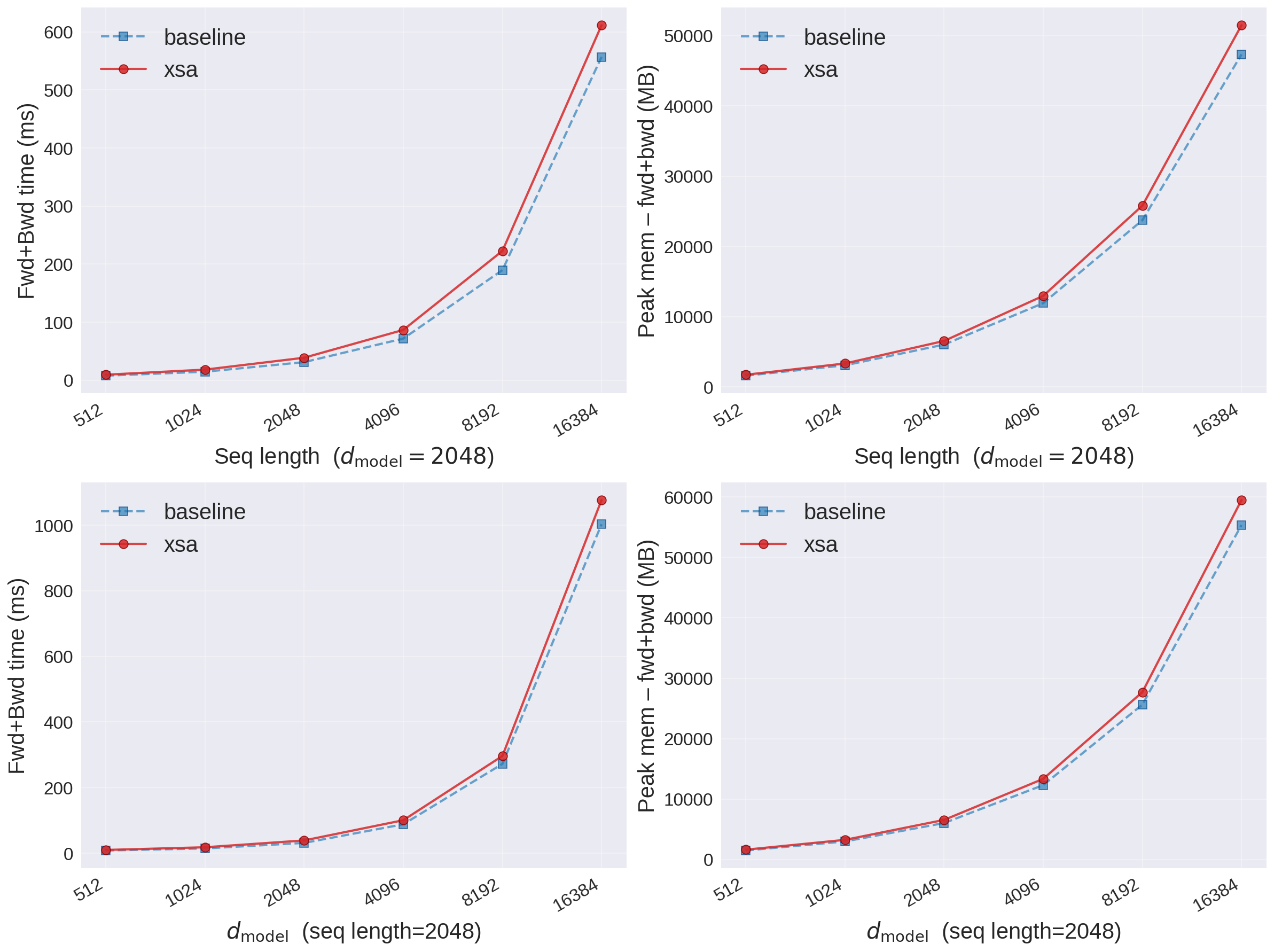}
    \caption{Time and memory efficiency of XSA compared to standard attention. XSA introduces minimal computational overhead across various sequence lengths and model sizes $d_{model}$.}
    \label{fig:speed_mem}
\end{figure}

\begin{table}[t]
\small
\caption{Architectures and learning rates. All models are trained with a batch size of $0.5M$ tokens, for a total of 200K iterations.}
\label{tab:model_sizes}
\centering
\renewcommand{\arraystretch}{1.2}
\begin{tabular}{lccccccc}
\toprule
Model Size
& $n_{\text{layers}}$
& $d_{\text{model}}$
& $n_{\text{heads}}$
& $d_{\text{head}}$
& Learning Rate \\
\midrule
0.7B
& 24
& 1536
& 6
& 256
& $5.0 \times 10^{-4}$ \\
\midrule
1.4B
& 24
& 2048
& 24
& 128
& $4.0 \times 10^{-4}$ \\
\midrule
2.7B
& 32
& 2560
& 24
& 128
& $3.0 \times 10^{-4}$ \\

\bottomrule
\end{tabular}
\end{table}

\begin{figure}[!h]
    \centering
    \includegraphics[width=0.49\linewidth]{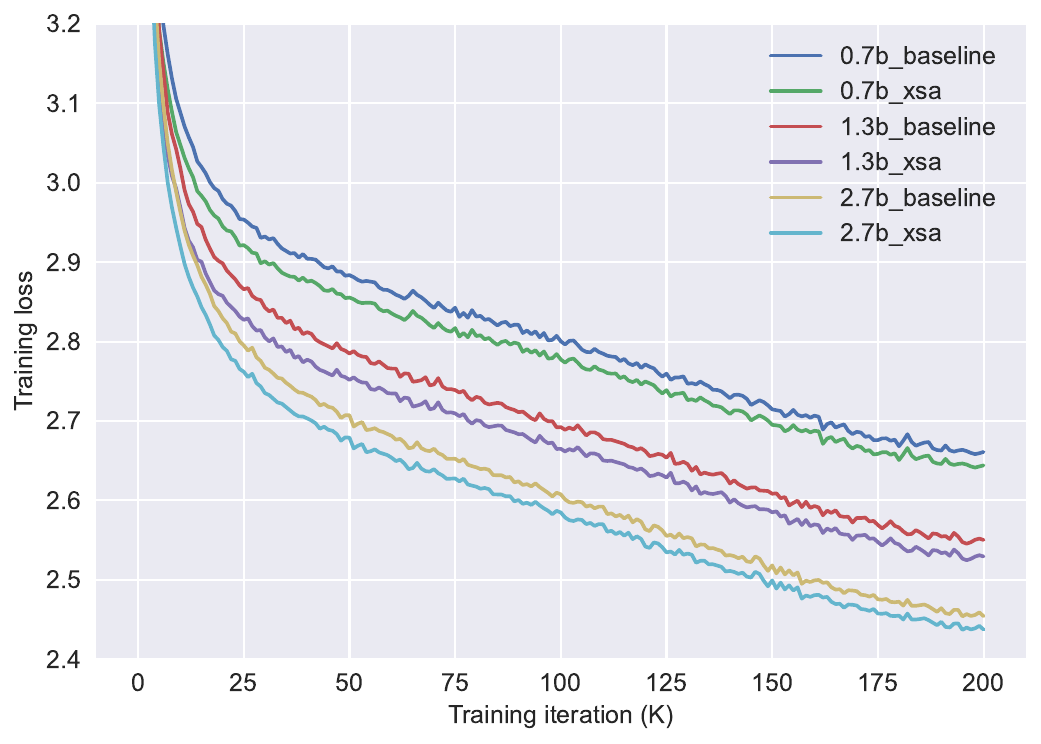}
    \includegraphics[width=0.49\linewidth]{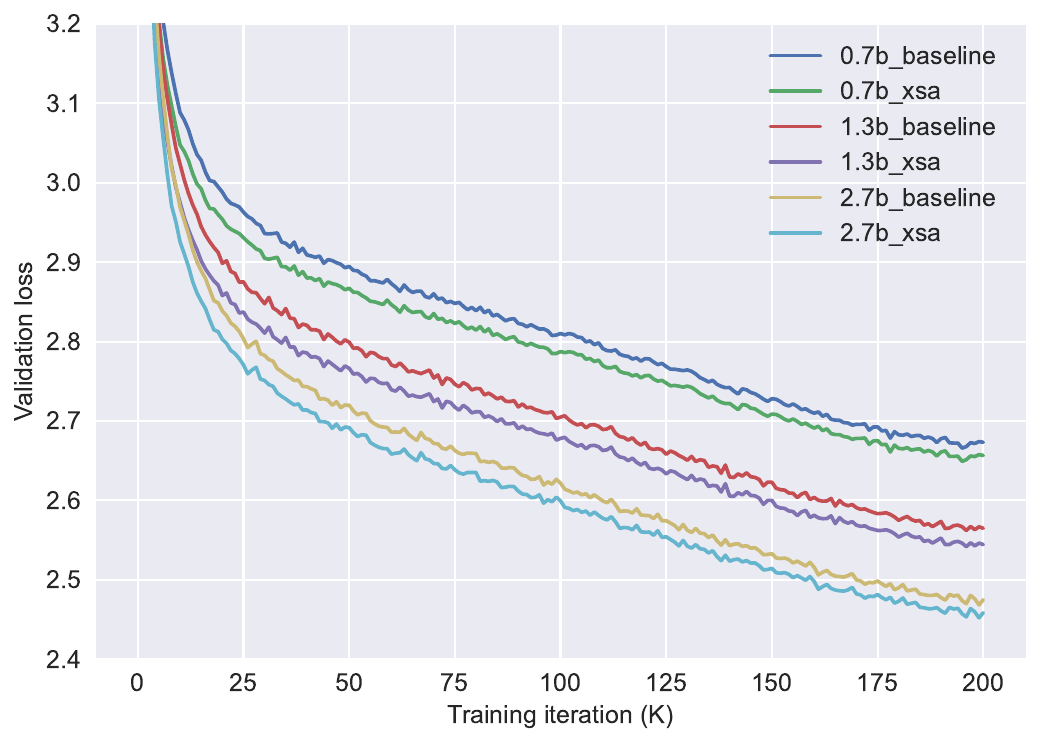}
    \caption{Training and validation loss curves of XSA against the baseline Transformer for three model sizes.}
    \label{fig:model_size}
\end{figure}

\subsection{Results}

\begin{table*}[!h]
\scriptsize
\centering
\caption{Downstream evaluation results of XSA vs baseline Transformer.}
\label{tab:main_results}
\begin{tabular}{
l
c
c
c
c
c
c
c
c
c
c 
}
\toprule
\textbf{Model} &
\textbf{ARC-E} &
\textbf{BoolQ} &
\textbf{HSwag} &
\textbf{LAMBADA} &
\textbf{OBQA} &
\textbf{PIQA} &
\textbf{SocIQA} &
\textbf{WinoGr} &
\textbf{Avg} &
\textbf{$\Delta$Avg} \\
\midrule
\multicolumn{11}{l}{\itshape \textbf{0.7B}} \\
Baseline & 51.26 & 61.07 & 55.68 & 52.82 & \textbf{35.00} & \textbf{74.05} & 40.02 & 55.88 & 53.22 & \multirow{2}{*}{\textbf{+0.26}}\\
XSA  & \textbf{52.69} & \textbf{61.19} & \textbf{56.29} & \textbf{54.07} & 32.20 & 73.78 & \textbf{41.45} & \textbf{56.20} & \textbf{53.48} & \\
\midrule
\multicolumn{11}{l}{\itshape \textbf{1.3B}} \\
Baseline & 56.19 & \textbf{65.47} & 60.69 & 56.24 & 34.60 & 75.90 & 41.40 & 58.80 & 56.16 & \multirow{2}{*}{\textbf{+1.03}}\\
XSA  & \textbf{58.84 }& 62.29 & \textbf{62.41} & \textbf{58.57} & \textbf{36.00} & \textbf{76.61} & \textbf{42.84} & \textbf{59.98} & \textbf{57.19}\\
\midrule
\multicolumn{11}{l}{\itshape \textbf{2.7B}} \\
Baseline & 58.59 & 60.98 & 66.20 & 60.18 & 37.00 & 76.61 & \textbf{42.94} & 61.96 & 58.06 &\multirow{2}{*}{\textbf{+1.36}}\\
XSA  & \textbf{60.65} & \textbf{64.86} & \textbf{67.40} & \textbf{62.04} & \textbf{38.40} & \textbf{77.80} & 41.45 & \textbf{62.75} & \textbf{59.42}\\
\bottomrule
\end{tabular}
\end{table*}

\begin{figure}[!b]
    \centering
    \includegraphics[width=0.49\linewidth]{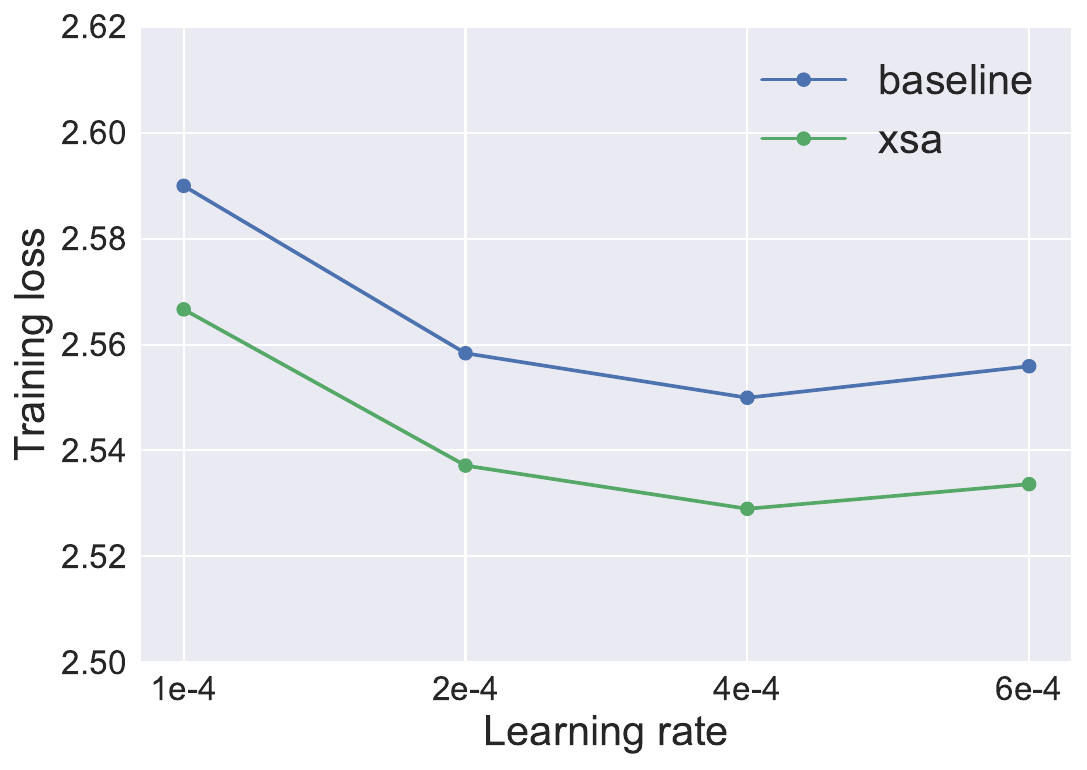}
    \includegraphics[width=0.49\linewidth]{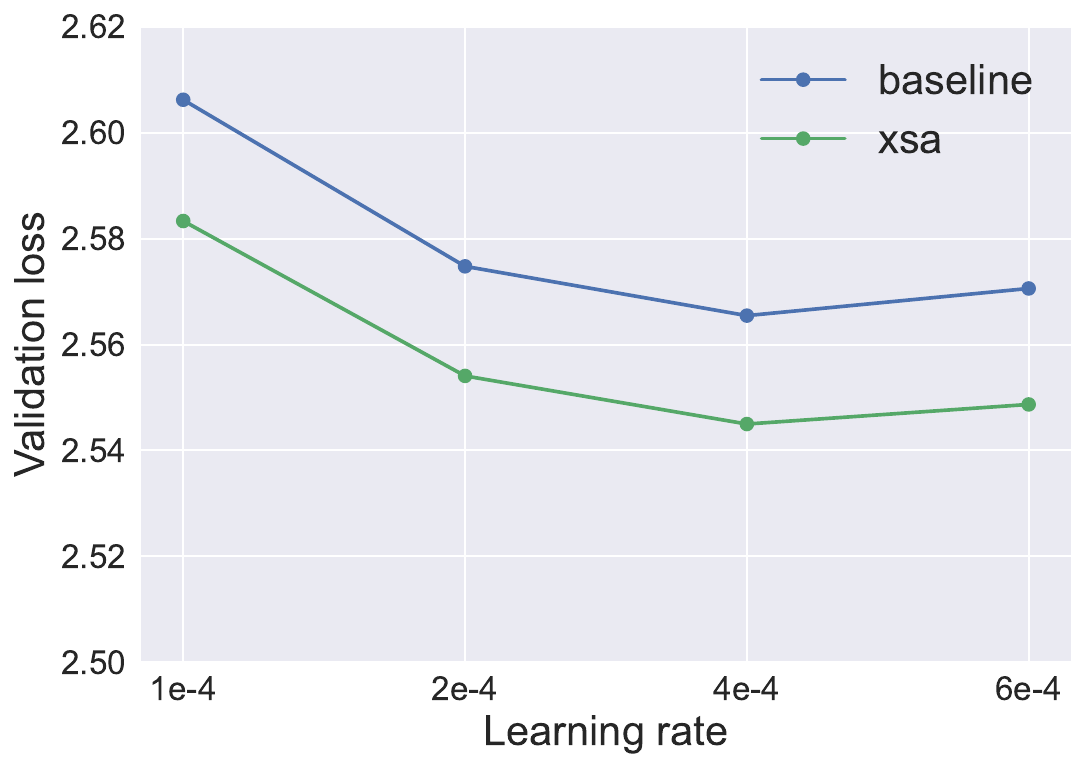}
    \caption{Training and validation loss of XSA against the baseline Transformer for various learning rates evaluated with the 1.3B model.}
    \label{fig:learning_rate}
\end{figure}

\begin{figure}[!b]
    \centering
    \includegraphics[width=0.49\linewidth]{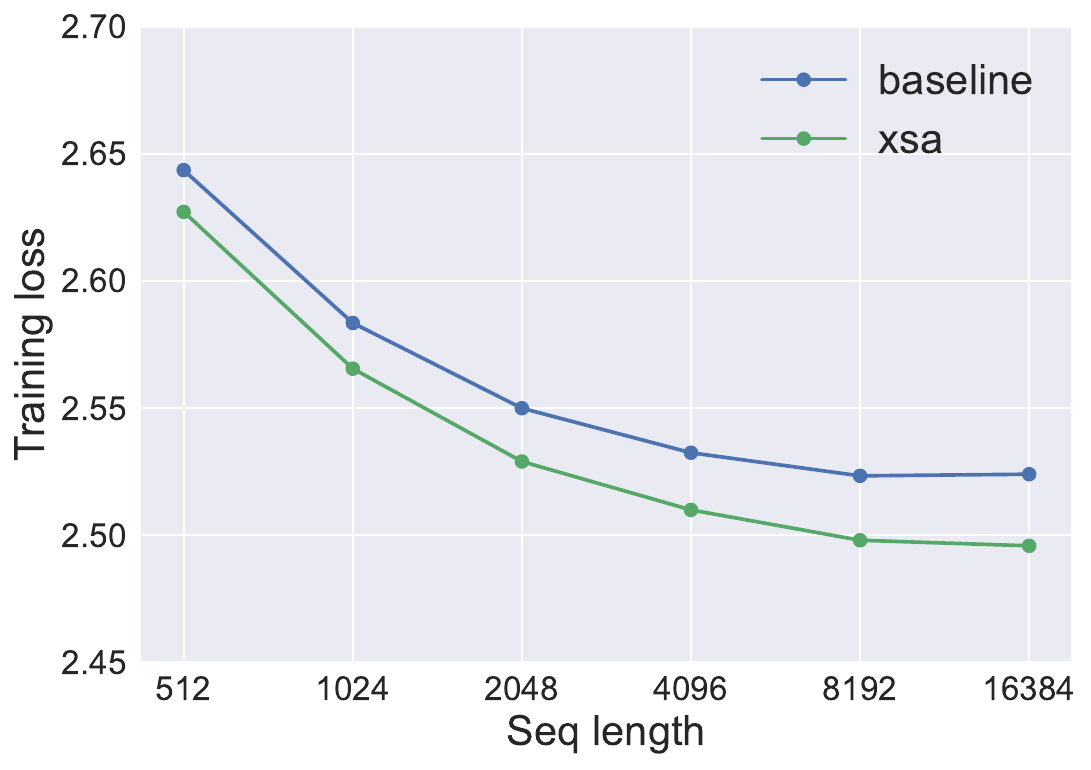}
    \includegraphics[width=0.49\linewidth]{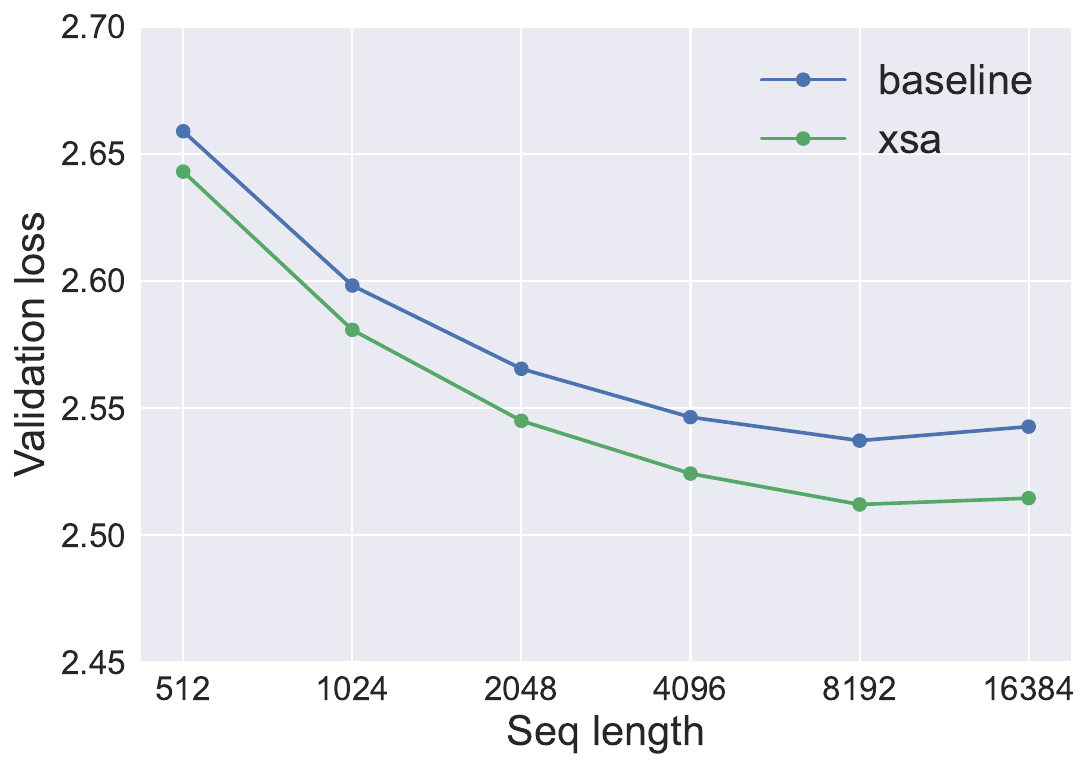}
    \caption{Training and validation loss of XSA against the baseline Transformer for various sequence lengths evaluated with the 1.3B model.}
    \label{fig:seq_len}
\end{figure}

\begin{figure}[!h]
    \centering
    \includegraphics[width=0.49\linewidth]{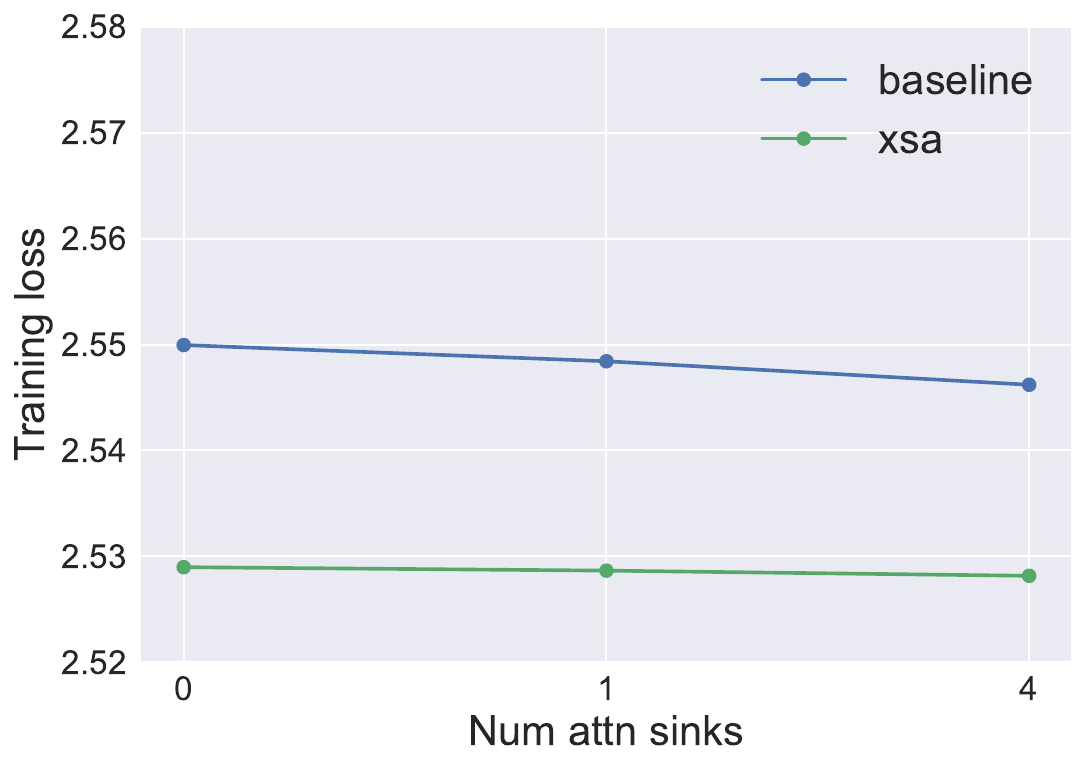}
    \includegraphics[width=0.49\linewidth]{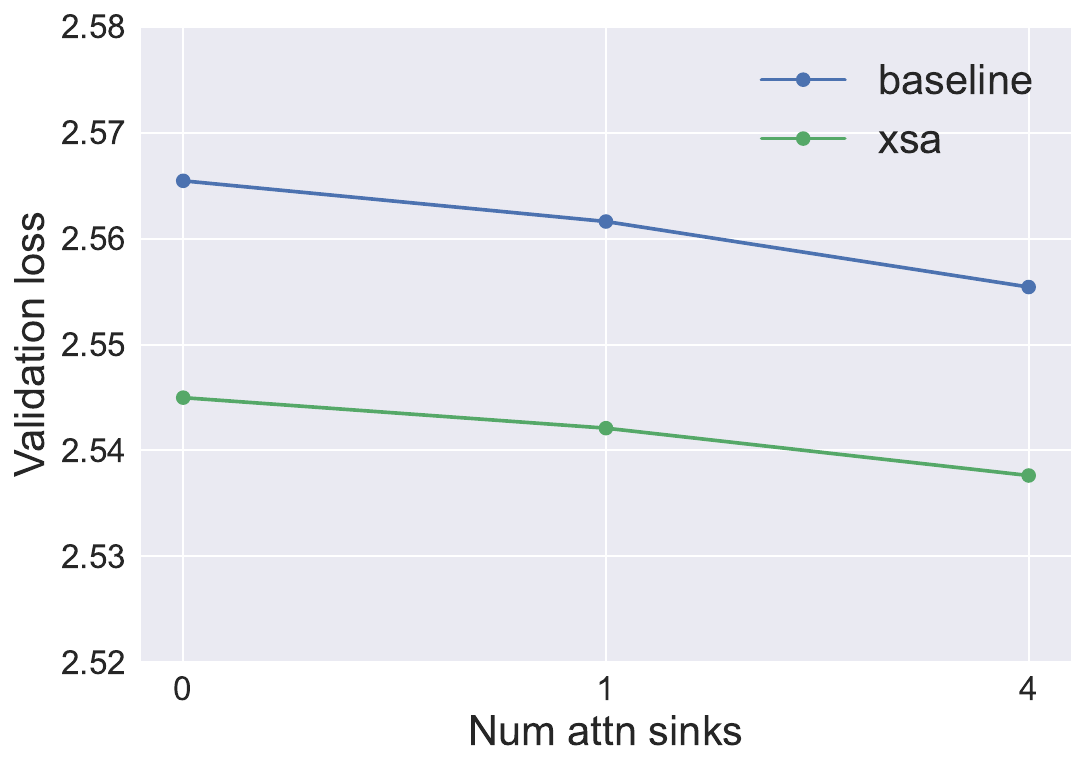}
    \caption{Training and validation loss of XSA against the baseline Transformer for various number of Attention Sinks evaluated with the 1.3B model.}
    \label{fig:attn_sink}
\end{figure}

\paragraph{Computational overhead} We first benchmark XSA in terms of its speed and memory efficiency. We run evaluations of an attention block (attention + FFN) on varied sequence lengths and model sizes, while switching XSA on and off. The experiments are ran on a B200 GPU with a batch size 32 and numerical precision of bfloat16. The results are shown in Figure~\ref{fig:speed_mem}. We see that, as expected, XSA introduces minimal overhead in terms of both speed and memory.

\paragraph{Model size} We next show the training and validation loss curves for the three model sizes in Figure~\ref{fig:model_size}. We observe that through the course of training XSA maintains a clear margin over the baseline in all three model sizes, across both training and validation. Furthermore, we conduct evaluations of the final checkpoints on 8 downstream tasks: ARC-Easy~\citep{clark2018think}, BoolQ~\citep{clark2018think}, HellaSwag~\citep{zellers2019hellaswag}, LAMBADA~\citep{paperno2016lambada}, OpenBookQA~\citep{mihaylov2018can}, PIQA~\citep{ba2016layer}, SocialIQA~\citep{sap2019social} and WinoGrande~\citep{sakaguchi2021winogrande}, which cover language, knowledge and reasoning aspects of the models. We run all evaluations with the Language Model Evaluation Harness~\citep{eval-harness}, and report the results w.r.t. accuracy (for BoolQ, LAMBADA, SocialIQA and WinoGrande) or length normalized accuracy (for ARC-Easy, HellaSwag, OpenBookQA and
PIQA) in Table~\ref{tab:main_results}. We see that across three model sizes, XSA consistently outperforms the baseline Transformer in terms of the average accuracy, with a larger margin as the model size increases.
Based on these observations, we speculate that XSA will remain advantageous in even larger scale training settings, both in terms of model size and training data size. 

\paragraph{Learning rate} It is also important to make sure that the performance gain of XSA holds across different learning rates. In order to test it, we select the 1.3B model and compare XSA with the baseline with four different learning rates. The comparison is shown in Figure~\ref{fig:learning_rate}. Again, we see that there is a near constant margin across all learning rates, demonstrating the robustness of the XSA architecture.

\paragraph{Sequence length} Next, we evaluate XSA's compatibility with long contexts. We again use the 1.3B model and train on six different sequence lengths, ranging in $\{512, 1024, 2048, 4096, 8192, 16384\}$. We use the same learning rate for all settings, and adjust the batch size such that the number of tokens per batch remains constant (0.5M). The result is shown in Figure~\ref{fig:seq_len}. Interestingly, XSA claims larger gains as sequence length increases. We suspect that this is due to the increasing tension on context modeling for longer sequences, which makes the benefit of XSA more pronounced. This also suggests that XSA is a promising technique for long context modeling, one of the critical problems of scaling Transformers. 

\paragraph{Comparison to Attention Sink} Incidentally, XSA is also related to Attention Sink~\citep{xiao2023efficient}. Instead of explicitly prepending to the sequence a set of learned sink tokens, XSA is able to allocate undesired attention scores to $a_{i,i}$. Therefore, XSA can be viewed as an implicit attention sink, and it would be interesting to compare it with standard Attention Sinks. We ran additional experiments by including learned attention sinks for both the baseline and XSA models, and report the results in Figure~\ref{fig:attn_sink}. We see that XSA maintains the loss margin in the existence of attention sinks.

\section{Discussions}
We have shown the exclusive self attention (XSA) demonstrates promising performance on standard language modeling tasks. However, many questions remain: How does it work at even larger scale w.r.t. model and data? Is it compatible with other optimizers such as Muon~\citep{jordan2024muon}? Does it work for other tasks/modalities besides language modeling? We hope that this study inspires future works that can properly answer these questions. 

\bibliographystyle{plainnat}
\bibliography{refs}

\end{document}